\documentclass[runningheads]{llncs}
\usepackage{graphicx}
\usepackage{booktabs}
\usepackage{adjustbox}
\usepackage[colorlinks=true,citecolor=blue,anchorcolor=purple]{hyperref}\usepackage{url}
\usepackage{multirow}
\usepackage{amsmath}
\usepackage{amssymb}
\usepackage{mathtools}
\usepackage{caption}
\usepackage{wrapfig}
\usepackage{lipsum}
\usepackage{amsfonts}
\usepackage{pifont}
\usepackage{fdsymbol}
\newcommand{\cmark}{\ding{51}}%
\newcommand{\xmark}{\ding{55}}%

\def \Rm{{\mathbb{R}}}

\def \gbf{{\mathbf g}}

\def \hbf{{\mathbf h}}

\def \mbf{{\mathbf m}}

\def \vbf{{\mathbf v}}

\def \Wbf{{\mathbf W}}

\def \Xbf{{\mathbf X}}

\def \Ybf{{\mathbf Y}}

\def \0bf{{\mathbf 0}}

\def \Lcal{{\mathcal L}}

\title{Operating critical machine learning models in resource constrained regimes}

 \author{Raghavendra Selvan\inst{1,2} \and Julian Schön\inst{1} \and Erik B Dam~\inst{1} }
 
\authorrunning{Selvan, Schön and Dam}
\titlerunning{Operating critical ML with resource constraints}
\institute{Department of Computer Science, University of Copenhagen \and
Department of Neuroscience, University of Copenhagen
\\
\email{raghav@di.ku.dk}
}

\begin{document}

\maketitle

\begin{abstract} 

	The accelerated development of machine learning methods, primarily deep learning, are causal to the recent breakthroughs in medical image analysis and computer aided intervention. The resource consumption of deep learning models in terms of amount of training data, compute and energy costs are known to be massive. These large resource costs can be barriers in deploying these models in clinics, globally. To address this, there are cogent efforts within the machine learning community to introduce notions of resource efficiency. For instance, using quantisation to alleviate memory consumption. While most of these methods are shown to reduce the resource utilisation, they could come at a cost in performance. In this work, we probe into the trade-off between resource consumption and performance, specifically, when dealing with models that are used in critical settings such as in clinics.\footnote{Source Code: \url{https://github.com/raghavian/redl}} 


\end{abstract}

\section{Introduction}
Every third person on the planet does not {\em yet} have universal health coverage according to the estimates from the United Nations\cite{sdg2022}. And improving access to universal health coverage is one of the UN Sustainable Development Goals. The use of machine learning (ML) based methods could be instrumental in achieving this objective. Reliable ML in clinical decision support could multiply the capabilities of healthcare professionals by reducing the time spent on several tedious and time-consuming steps. This can free up the precious time and effort of healthcare professionals to cater to the people in need\cite{yu2018artificial}. 

The deployment of ML in critical settings like clinical environments is, however, currently in a nascent state. In addition to the fundamental challenges related to multi-site/multi-vendor validation, important issues pertaining to fairness, bias and ethics of these methods are still being investigated\cite{ricci2022addressing}. The demand for expensive material resources, in terms of computational infrastructure and energy requirements along with the climate impact of ML are also points of concern\cite{selvan2022carbon}.

In this work, we focus on the question of improving the resource efficiency of developing and deploying deep learning models for resource constrained environments~\cite{sze2017efficient,bartoldson2022compute}. Resource constraints in real world scenarios could be manifested as the need for: low latency (corresponding to urgency), non-specialized hardware (instead of requiring GPU/similar infrastructure) and low-energy requirements (if models are deployed on portable settings). The case for resource efficiency when {\em deploying} ML models is quite evident due to these aforementioned factors. In this work, we {\em also} argue that the training resource efficiency is important, as ML systems will be most useful in the clinical settings if they are able to continually learn~\cite{baweja2018towards,vokinger2021continual}. We validate the usefulness of several methods for improving resource efficiency of deep learning models using detailed experiments. We show that resource efficiency -- in terms of reducing compute time, GPU memory usage, energy consumption -- can be achieved without degradation in performance for some classes of methods.


\section{Methods for Resource Efficiency}
\label{sec:methods}
In this work we mainly consider reducing the overall memory footprint of deep learning models, which in turn could reduce computation time and the corresponding energy costs. We do not explicitly study the influence of model selection using efficient neural architecture search~\cite{tan2019efficientnet} or model compression using pruning or similar methods~\cite{cheng2020compression}, as these techniques require additional design choices outside of the normal model development and deployment phases. By focusing on strategies that can be easily incorporated, we also place emphasis on the possible ease of clinical adoption.

Denote a neural network, $f_\Wbf(\cdot): \Xbf \rightarrow \Ybf$ which acts on inputs $\Xbf$ to predict the output $\Ybf$ and has trainable parameters $\Wbf$. For recent deep neural networks, the number of trainable parameters, $|\Wbf|$, can be quite large~\cite{sevilla2022compute} in $\mathcal{O}(100M)$. 

During training, the weights, $\Wbf$, are updated using some form of a gradient-based update rule. This requires the computation of the gradients of the loss $\Lcal(\cdot)$ with respect to $\Wbf$ at iteration $t$, $\gbf_t=\frac{\partial{\Lcal}}{\partial{\Wbf}}$. In the case of stateful optimisers that use momentum-based optimisation, such as Adam~\cite{kingma2014adam}, the first and second order statistics of the gradient over time, $\mbf_t,\vbf_t$, respectively, are maintained for improved convergence. Thus, for a neural network with $|\Wbf|$ parameters, at any point during the training there are an additional $\approx 3\cdot |\Wbf|$ variables stored in memory. Further, the intermediate activations at layer $l$, $\hbf_l$ are also stored in memory to efficiently perform backpropagation. Note that all the scalar entries of $\Wbf, \gbf_t, \mbf_t, \vbf_t, \hbf_l \in \Rm$. On most computers, these real numbers are discretised into floating point-32 (FP32) format; wherein each variable requires 32 bits.

Resource efficiency in this work is primarily addressed by reducing the precision of these variables by quantisation~\cite{fiesler1990weight,sze2017efficient}. The implication of using quantised or low-precision variables is that it not only reduces memory footprint but can also reduce the compute cost of matrix multiplications~\cite{hubara2017quantized} at the expense of loss in precision. Note, however, that the overhead of performing quantisation in some cases might outweigh the gains in computation.

In this work, we investigate a combination of the following three quantisation strategies:
\begin{enumerate}
    \item {\bf Gradients and intermediate activations}: Drastic quantisation of $\hbf_l, \gbf_t$ have been studied extensively in  literature~\cite{hubara2017quantized,chakrabarti2019backprop}. For instance, in~\cite{hubara2017quantized} the gradients and activations were quantised to 1-bit and still yielded reasonable performance compared to the 32-bit models. However, not all operations are well suited for quantisation. In frameworks like Pytorch~\cite{paszke2019pytorch}, the precision of $\hbf_l, \gbf_t$ are automatically cast into 16-bit precision, and recast to 32-bit when necessary. This technique known as {\tt automatic mixed precision} (AMP) can yield considerable reduction of memory costs~\cite{micikevicius2018mixed}.
    \item {\bf Optimiser states}: Until recently, the main focus of quantisation in deep learning has been on compressing $\gbf_t,\hbf_l,\Wbf$. In a recent work in~\cite{dettmers8}, it was shown that the optimiser states, $\mbf_t,\vbf_t$ can use up to 75\% of memory. By quantising optimiser states to 8-bits using a dynamic quantisation scheme, this work reduces the memory consumption during model training. 
    \item {\bf Model weights}: The resolution of the hypothesis space is controlled by the precision of the weights, $\Wbf$. Using lower precision weights for models can reduce their expressive power. This is most evident when a model trained in full precision is then cast into lower precision. In such cases, it is a common strategy to fine tune the quantised model for a few epochs to recover the loss in performance~\cite{nagel2021white}. However, when a model is trained from the outset in low-precision modes, they can achieve similar performances to models with full precision weights~\cite{sun2020ultra}.
\end{enumerate}

\section{Data \& Experiments}

\begin{figure}[t]

\begin{minipage}{0.55\textwidth}
    \centering
\includegraphics[width=0.79\textwidth]{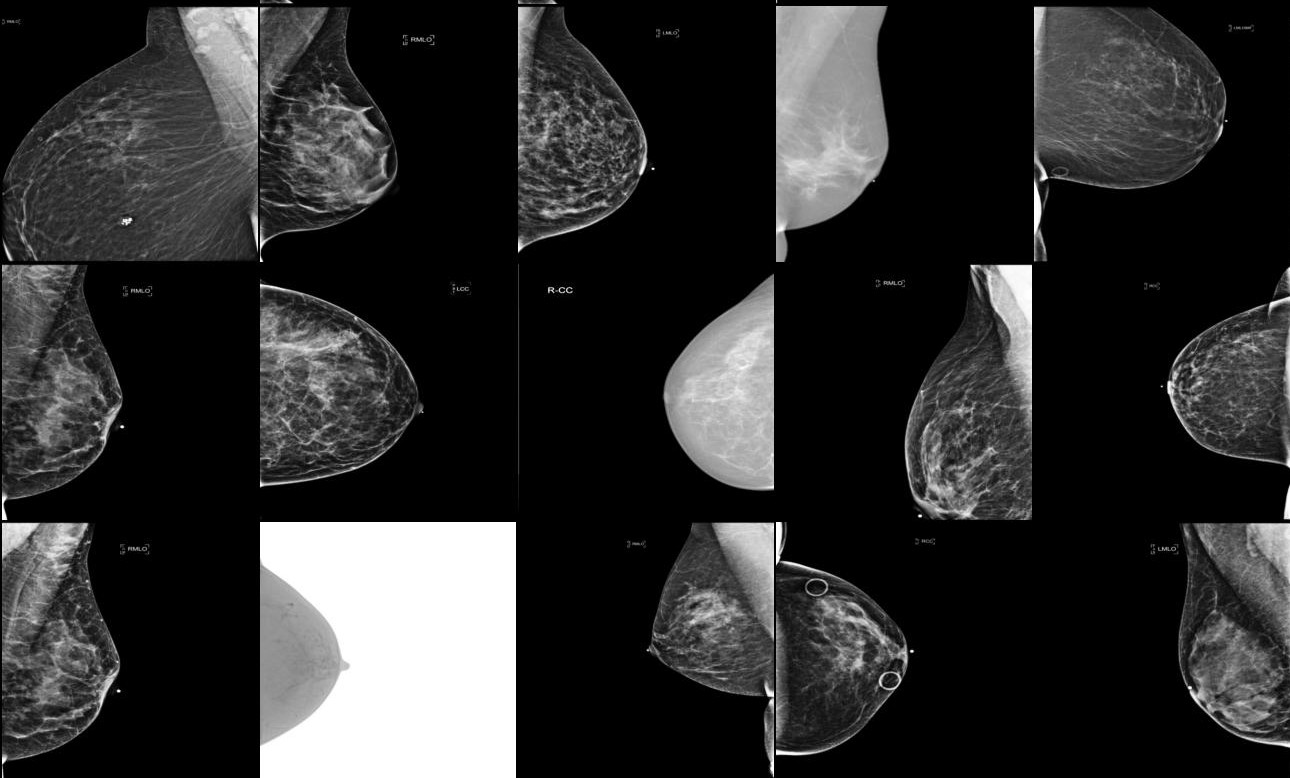}

(A)
\end{minipage}
\begin{minipage}{0.4\textwidth}
    \centering
    \includegraphics[width=0.65\textwidth]{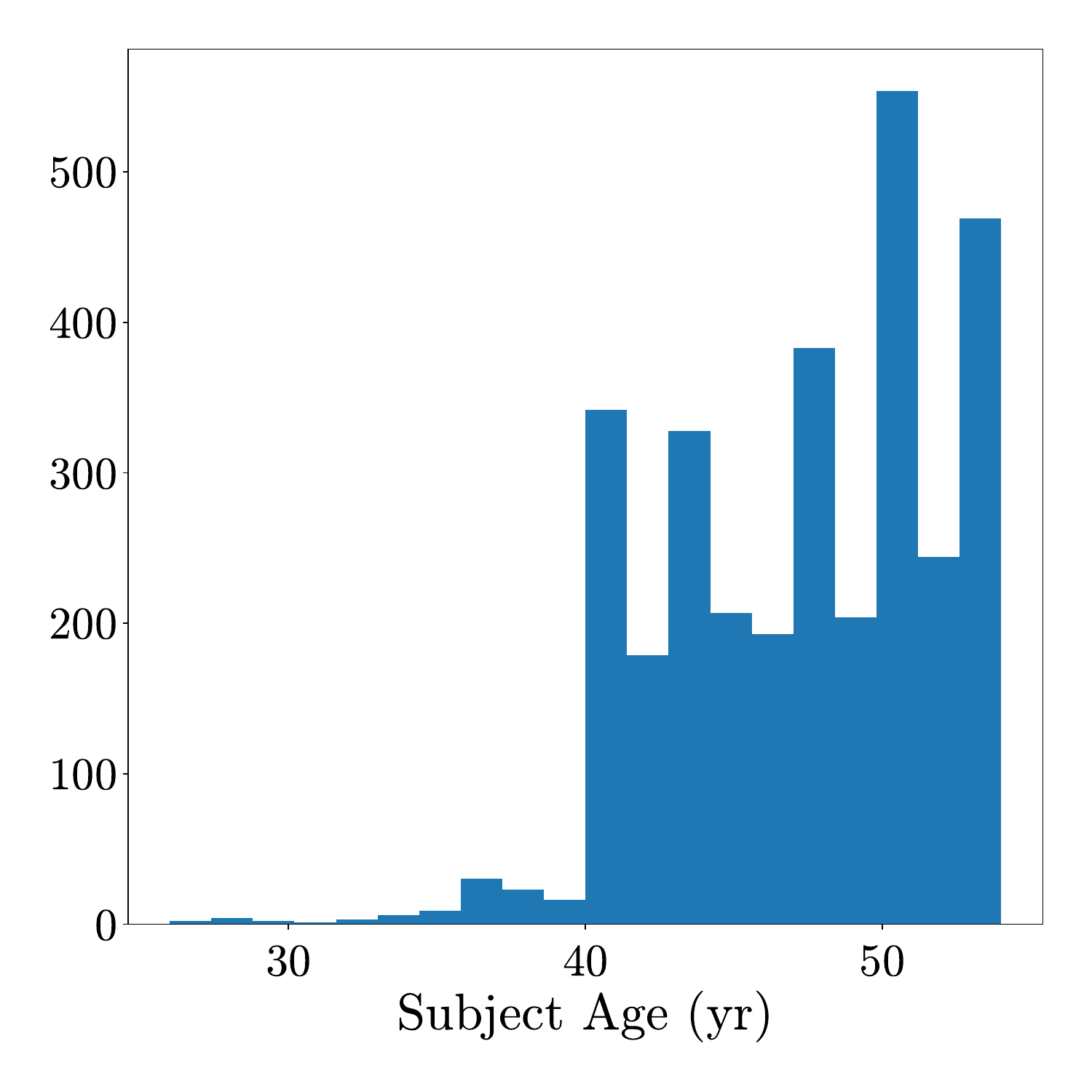}
    
    (B)
\end{minipage}
    \vspace{-0.35cm}
\caption{(A): Sample mammography images from the RSNA training set showing the diverse acquisition and anatomical variations. (B): Age distribution of the subjects in the dataset used. Maximum age of $55$ yr was used. }
    \label{fig:data}
    \vspace{-0.5cm}
\end{figure}

Nation-wide screening for early detection of cancer is being attempted in many countries. For instance, early breast cancer detection using mammography screening has shown positive impact, both for the subjects and in allocating healthcare resources~\cite{daysal2022economic}. To simulate a scenario where ML models are deployed in such screening programs, we use a publicly available mammography screening dataset for the task of recommending follow-up scans. We study the influence of using resource optimised models and their impact on the predictions in the mammography screening.
\\
{\bf Datasets}: The experiments in this work were primarily performed using a subset of the data available as part of the Radiological Society of North America (RSNA) Mammography Breast Cancer Detection challenge\cite{rsna-breast-cancer-detection}\footnote{\url{https://www.kaggle.com/competitions/rsna-breast-cancer-detection/overview}. Accessed on 08/03/2023}. Subjects under the age of $55$ yr in the training set were used as the cohort of interest for early detection. This yielded a dataset comprising $3199$ unique subjects. As each subject has multiple scans (also from different views), the final dataset consists of $11856$ mammography scans. While the dataset provides multiple markers which could be used as labels, we focus on the BI-RADS\footnote{BI-RADS:Breast Imaging Reporting \& Data System} score\cite{orel1999birads}. The dataset provides three grades of the BI-RADS score: 0 if the breast required follow-up, 1 if the breast was rated as negative for cancer, and 2 if the breast was rated as normal. We combine the classes 1 and 2 to make a binary classification task of predicting if a follow-up scan was required or not. This resulted in the following label distribution: [$7903$, $3953$], corresponding to requiring follow-up (class-1) and no follow-up (class-0). The dataset was randomly split into training, validation and test sets based on subject identities. This resulted in a split of [1919,640,640] subjects in the three sets, respectively. In terms of mammography images, the resulting split was [7113,2371,2372] for the training, validation and test sets. Sample images from the dataset and the age distribution of the subjects is shown in Figure~\ref{fig:data}. We acknowledge that this classification task is a simplification of the mammography reading done during screening, but depending on the screening setup (e.g. number of readers), even this simple task could supplement the workflow. 

Additional experiments on a second dataset for lung nodule detection are also reported for further validation. We use the LIDC-IDRI dataset which consists of 1018 thoracic CT scans and predict the presence or absence of tumours~\cite{armato2004lung}. Further details of the LIDC classification dataset are provided in the Supplementary Material.
\\
{\bf Classification models}: To evaluate the impact of resource constraints, different classes of deep learning models are used. Specifically, we use Densenet~\cite{huang2017densely}, Efficientnet~\cite{tan2019efficientnet}, Vision transformer~\cite{dosovitskiyimage} and Swin Transformer~\cite{liu2021swin}. All models were pretrained on ImageNet and fine tuned on the two datasets\footnote{Pretrained weights for the models were obtained from \url{https://timm.fast.ai/}}. 
\\
{\bf Experimental design}: All the models were trained for a maximum of 50 epochs, with an early stopping patience of 5 epochs using batch size of 32. All the experiments were performed on a desktop workstation with Intel i7 processor, 32GB memory and Nvidia RTX 3090 GPU with 24 GB memory. All experiments were performed in
Pytorch (v1.7.1)~\cite{paszke2019pytorch}, and unless otherwise specified using the Adam~\cite{kingma2014adam} optimizer with learning rate of $10^{-5}$. Measurements of run time, energy and carbon consumption were performed using CarbonTracker~\cite{anthony2020carbontracker}. All models were trained three times with different random seeds. The training of models in this work used 25.0 kWh of electricity contributing to 4.7 kg of CO2eq.
%
\\
{\bf Experiments}: Resource constraints in real world scenarios could be due to: compute time (corresponding to low latency), specialized hardware (access to GPU infrastructure) and energy consumption (if models are deployed on portable devices). Using these and the test set performance as the axes, we run all the classification models with- and without- any resource optimisation. The three main optimisation techniques we use are related to the methods described in Section~\ref{sec:methods}. Firstly, we perform automatic mixed precision training of the models in order to obtain low compute and memory models~\cite{micikevicius2018mixed}. This is implemented using the {\tt automatic mixed precision} (AMP) feature in Pytorch. Secondly, we minimize the resource utilisation of the optimizer using an {\tt 8-bit optimiser} based on the implementation\footnote{8-bit optimiser from \href{https://github.com/TimDettmers/bitsandbytes}{bitsandbytes}} from~\cite{dettmers8}. Finally, we also investigate training of a low precision model by casting all the model weights to {\tt half precision} (float-16). We then compare the performance of all the classification models without any explicit resource optimisation and with the combination of the mixed precision training/half-precision models and the 8-bit optimiser. Note that AMP cannot be combined with Half Precision models, as the model itself is in low precision. We use binary accuracy on the test set ($P_T$), average training time to convergence ($T_c$), maximum GPU memory required, average inference time on test set ($T$) and the total energy consumption ($E$)  as the measures to compare the models and different resource efficiency techniques.
\\
\begin{table}[h]
\vspace{-0.5cm}
\caption{Quantitative comparison of the different classification methods reported over three random initialisations for RSNA and LIDC datasets. The use of 8-bit optimizer (8bit), automatic mixed precision (AMP), and half precision model (Half) are marked. Number of parameters: $|\Wbf|$, mean accuracy with standard deviation over three seeds, average GPU memory required (in GB), average inference time, and average energy consumption of different settings reported when predicting on test set. Models that diverged at least during one of the random initialisations are marked with $^*$ in $P_T$ column.}
\vspace{0.1cm}
\label{tab:results}
\centering
\tiny
\begin{tabular}{clrccccrrrr}
\toprule
        {\bf Dataset} &{\bf Method} &   $|\Wbf|$(M) &  {\bf 8bit} &  {\bf AMP} &   {\bf Half} & {$\mathbf P_T$}  &    {\bf GPU} &     $\mathbf T_{c}$(s) & {\bf T}(s) & {\bf E}(Wh) \\
\midrule
\multirow{16.5}{*}{\bf RSNA\cite{rsna-breast-cancer-detection}}&\multirow{5}{*}{Densenet\cite{huang2017densely}} &   \multirow{5}{*}{6.9}&    \xmark &    \xmark & \xmark & 0.738$\pm$0.005 &    6.6 &  361.3 & 2.7 &  104.2 \\
    &&&    \xmark &    \cmark & \xmark & 0.740$\pm$0.005 &  3.3 &  296.0 & 2.7 &  59.2 \\
    &&&    \cmark &    \xmark & \xmark &0.743$\pm$0.003 &  6.5 &  240.0 & 2.7 & 58.1 \\
    &&&    \cmark &    \xmark & \cmark & {\bf 0.745$\pm$0.004} &  
    {\bf 3.2} &  {\bf 170.3} & {\bf 2.1} & 34.6\\
    &&&    \cmark &    \cmark & \xmark & {\bf 0.744$\pm$0.006} &  
    3.2 &  191.3 & 2.7&  42.3 \\
            \cmidrule{2-11}
 &       \multirow{5}{*}{Swin Trans.\cite{liu2021swin}} & \multirow{5}{*}{86.9}&   \xmark &    \xmark & \xmark & 0.739$\pm$0.008  & 15.3 &  617.6 & 7.5 & 100.1 \\
         &&&    \xmark &    \cmark & \xmark & 0.735$\pm$0.006  & 10.7 &  497.0 & 7.5& 109.8 \\
         &&&    \cmark &    \xmark & \xmark & 0.739$\pm$0.012  & 14.3 &  732.3 & 7.5& 86.1 \\
         &&&    \cmark &    \xmark & \cmark & 0.733$\pm$0.013  &  
         7.5 &  316.0 & 4.3& 118.5 \\ 
         &&&    \cmark &    \cmark & \xmark & 0.712$\pm$0.01$^*$  &  
         9.7 &  588.6 & 7.5&  95.8 \\ 
         \cmidrule{2-11}
 &         \multirow{5}{*}{ViT\cite{dosovitskiyimage}} &  \multirow{5}{*}{116.7}&     \xmark &    \xmark & \xmark & 0.728$\pm$0.001  &  5.6 &  322.3 & 1.9 & 52.0 \\
            &&&    \xmark &    \cmark & \xmark & 0.728$\pm$0.006  &  5.6 &  249.0 &1.9 & 39.8 \\
            &&&    \cmark &    \xmark & \xmark & 0.716$\pm$0.009  &  4.1 &  497.0 & 1.9& 81.0 \\
            &&&    \cmark &    \xmark & \cmark & 0.690$\pm$0.02$^*$  
            &  3.5 &  {69.6} & { 1.0} &10.7 \\
            &&&    \cmark &    \cmark & \xmark & 0.665$\pm$0.02$^*$  
            &  4.7 &  132.6 & 1.9& 21.0 \\    
\midrule

\midrule
\multirow{16.5}{*}{\bf LIDC\cite{armato2004lung}}&\multirow{5}{*}{Densenet\cite{huang2017densely}} &   \multirow{5}{*}{6.9}&    \xmark &    \xmark & \xmark & 0.656$\pm$0.005 &    6.6 &  674.3 & 3.4 &  104.2 \\
    &&&    \xmark &    \cmark & \xmark & 0.655$\pm$0.010 &  3.3 &  425.7 & 3.4 &  59.2 \\
    &&&    \cmark &    \xmark & \xmark &0.677$\pm$0.008 &  6.5 &  370.7 & 3.4 & 58.1 \\
    &&&    \cmark &    \xmark & \cmark & { 0.675$\pm$0.007} &  
    {\bf 3.2} &  {\bf 234.3} & { 2.7} & 34.6\\
    &&&    \cmark &    \cmark & \xmark & {0.676$\pm$0.005} &  
    3.2 &  291.3 & 2.7&  42.3 \\
            \cmidrule{2-11}
 &       \multirow{5}{*}{Swin Trans.\cite{liu2021swin}} & \multirow{5}{*}{86.9}&   \xmark &    \xmark & \xmark & 0.697$\pm$0.018  & 15.3 &  678.7 & 9.6 & 109.8 \\
         &&&    \xmark &    \cmark & \xmark & 0.696$\pm$0.018  & 10.8 &  544.0 & 9.6& 86.1 \\
         &&&    \cmark &    \xmark & \xmark & 0.684$\pm$0.016  & 14.4 &  727.3 & 9.6&118.5 \\
         &&&    \cmark &    \xmark & \cmark & {\bf 0.704$\pm$0.007}  &  9.9 &  401.0 & 5.6&  64.1 \\ 
         &&&    \cmark &    \cmark & \xmark & 0.684$\pm$0.020  &  7.5 &  615.3 & 9.6& 99.9 \\ 
         \cmidrule{2-11}
 &         \multirow{5}{*}{ViT\cite{dosovitskiyimage}} &  \multirow{5}{*}{116.7}&     \xmark &    \xmark & \xmark & 0.610$\pm$0.041  &  5.6 &  231.7 & 2.4 & 37.2 \\
            &&&    \xmark &    \cmark & \xmark & 0.608$\pm$0.043  &  5.5 &  166.3 &2.4 & 25.8 \\
            &&&    \cmark &    \xmark & \xmark & 0.636$\pm$0.012  &  4.3 &  415.3 & 2.4& 67.1 \\
            &&&    \cmark &    \xmark & \cmark & 0.564$\pm$0.072  &  3.4 &  123.0 & {\bf 1.3}& 15.2 \\
            &&&    \cmark &    \cmark & \xmark & 0.561$\pm$0.076  &  4.5 &  {99.0} & 2.4 &19.1 \\
\bottomrule
\end{tabular}
\vspace{-0.5cm}
\end{table}
{\bf Results}: Experiments with the various resource optimisation settings and methods for both datasets are reported comprehensively in Table~\ref{tab:results}. For each method, five settings are reported. We treat the model with no additional resource optimisation as the {\em baseline}, reported in the first row for each method. For each model, we then report the following settings: [AMP, 8bit optimizer, AMP+ 8bit optimizer, Half precision + 8bit optimiser] yielding five settings for each model. The models are sorted based on the total number of trainable parameters, reported in the third column. 

At the outset, it is worth noting that the baseline model performance on the test set of all Densenet and the transformer-based Swin- and Vision- Transformers are comparable ($\approx 0.74$) in the RSNA mammography dataset. 

The use of AMP during training does not degrade the performance of any of the models, while reducing the maximum GPU memory utilised (up to 50\% in most cases). The run time, $T_c$ for all the models is also decreased when using AMP.

The use of 8-bit optimiser reduces the GPU memory utilised and the convergence time. The more interesting observation is that in almost all cases (except ViT), it also converges to a better solution, yielding a small performance improvement. This is consistent with the observations reported in~\cite{dettmers8}, where this behaviour is attributed to the elimination of outliers (which otherwise could lead to exploding gradients) during quantisation and dequantisation of the optimiser states.

Setting the model weights to half precision (16-bit instead of 32-bit) immediately gives a reduction in the GPU memory utilisation. In our experiments, however, there were stability issues when the half precision model was trained using the standard Adam optimiser. Using the 8-bit optimiser which is quantisation-aware helped convergence~\cite{dettmers8}. 

The best performing configuration across all the models for RSNA dataset is the Densenet with 8-bit optimiser either when used with AMP or the half precision model. This is an interesting result showing that these techniques not only reduce resource utilisation but could also improve model convergence.

The observations about the influence of AMP, 8-bit optimiser and half-precision hold within each model category for the LIDC dataset. The best performing configuration across all the models for LIDC dataset is the Swin Transformer with 8-bit optimiser when used with the half precision model.

\begin{figure}[t]
\begin{minipage}{0.32\textwidth}
    \centering
    \includegraphics[width=0.99\textwidth]{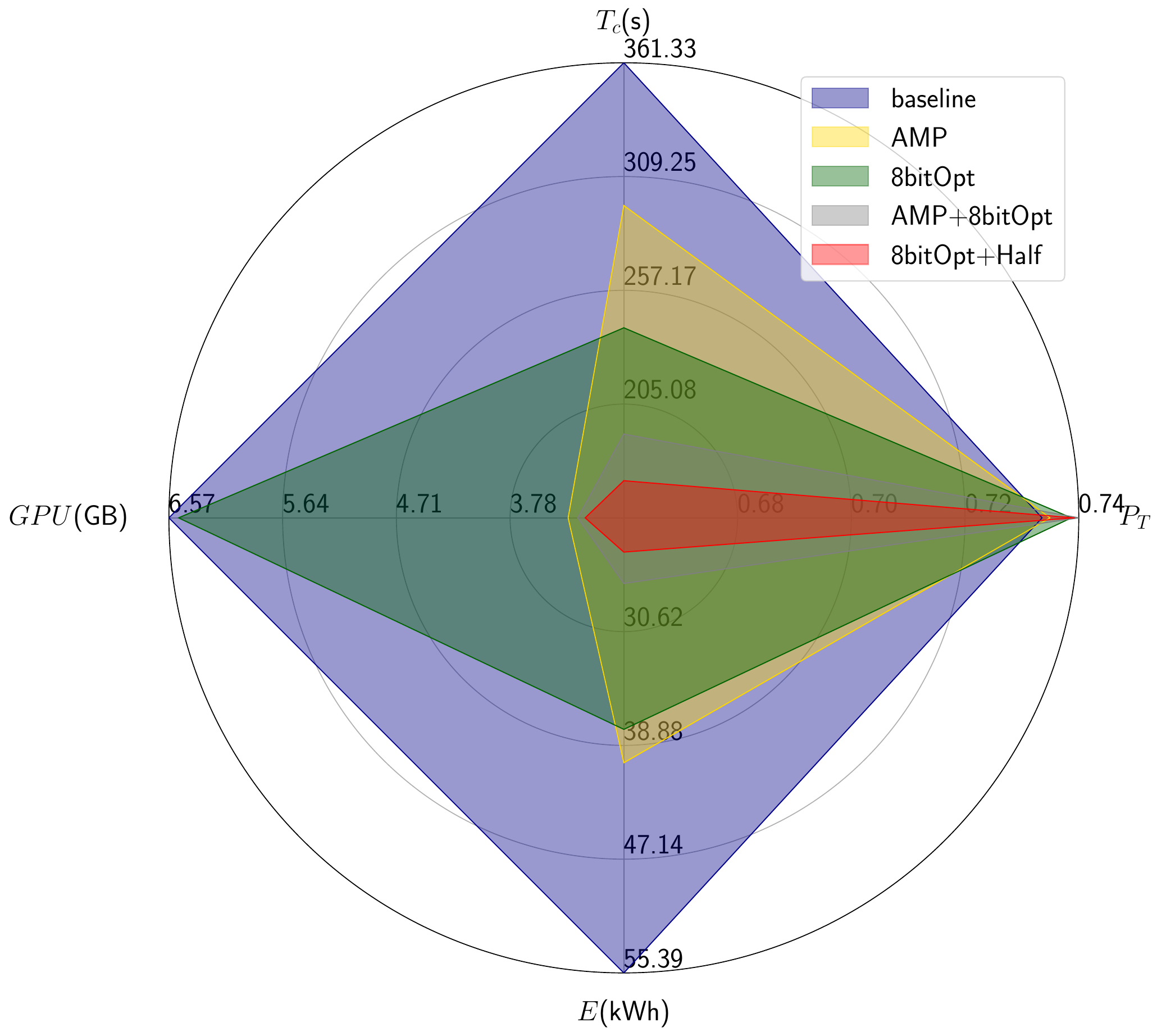}

(A): Densenet
\end{minipage}
\begin{minipage}{0.32\textwidth}
    \centering
    \includegraphics[width=0.99\textwidth]{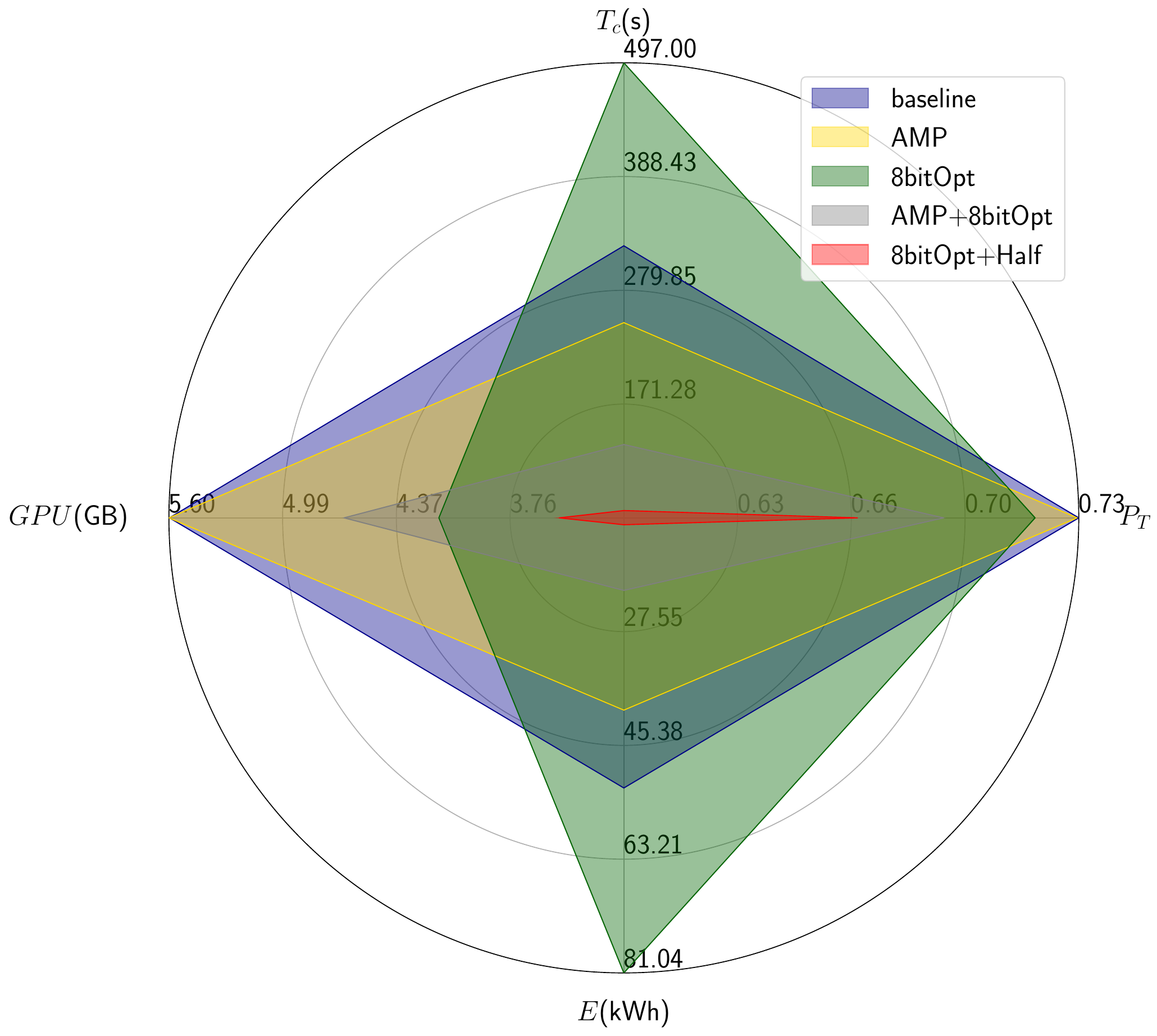}
    
    (B): ViT
\end{minipage}
\begin{minipage}{0.3\textwidth}
    \centering
    \includegraphics[width=0.95\textwidth]{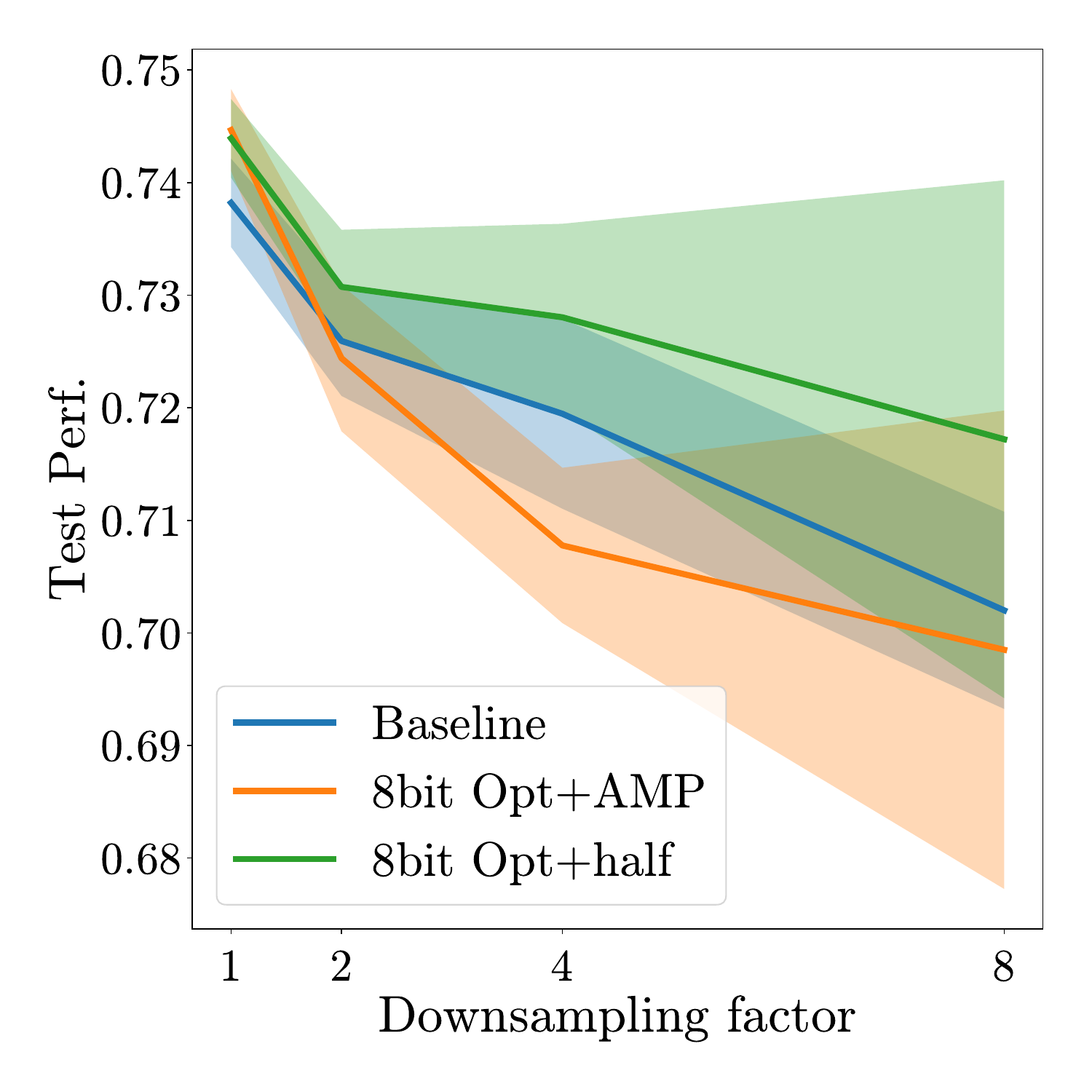}
    
    (C)
\end{minipage}
\caption{(A \& B): Radar plots showing the mean metrics for performance ($P_T, E, \text{GPU},T$) reported in Table~\ref{tab:results} for the five settings, shown for Densenet~\cite{huang2017densely} and Vision transformer~\cite{dosovitskiyimage}. (C): Influence of different scales of downsampling on the test performance for three configurations of Densenet models on the RSNA dataset. For each, the mean and the standard deviation of the three runs is shown as the curve and the shaded area, respectively.}
    \label{fig:results}
    \vspace{-0.5cm}
\end{figure}

The performance trends in Table~\ref{tab:results} are captured visually in the radar plots presented in Figure~\ref{fig:results}-A for Densenet. The mean performance measures of Densenet over the three random initialisations, for the five settings, are shown in different colours. The four axes are used to report four key measures:[$P_T, E, \text{GPU},T$]. The optimal model would have larger footprint along $P_T$ and smaller values on the other axes. We notice that the baseline model (in blue) spans a larger region over all axes, wheres the half precision model trained with 8-bit optimiser (in red) has smaller span. These trends are also visualised for the Vision transformer model in Figure~\ref{fig:results}-B, and for other models in the Supplementary Material.

\vspace{-0.25cm}
\section{Discussion}
\vspace{-0.25cm}
{\bf Data Quality}: One common method to manage the resource constraints during training or inference of deep learning models is the use of reduced quality data. For instance, in image analysis, downsampling of images is a commonly used strategy as this can have a direct impact on the size of the models. Such reduction in data quality are undoubtedly lossy. We illustrate this for the mammography screening task for the Densenet models in Figure~\ref{fig:results}-C. We reduce the quality of the images both during training and inference by downsampling the image at different scales: $S=[2,4,8]$. We observe a clear performance degradation compared to the models trained with full resolution ($S=1$). While this might be unsurprising, we argue, in lieu of the results in Table~\ref{tab:results} that optimising models instead of degrading data quality can offer a better trade-off between performance and resource efficiency.
\\
{\bf Resource Efficiency and Model Complexity}: The most common measure of model complexity is the number of parameters. The experiments and results reported in Table~\ref{tab:results} reveal slightly surprising trends. The best performing model for the RSNA dataset (Densenet) had the fewest number of parameters (6.9M), and ViT which was the largest model (116.7M) was also the fastest at inference time ($<$2s). This is due to the fact that during training the resource overhead for deep learning models is due to factors other than model weights, such as the optimiser states or other derived data like gradients/activation maps as discussed in Section~\ref{sec:methods}. We have demonstrated that resource efficiency can also be achieved by optimising these other axes.
\\
%
%
{\bf Resource Efficiency of Vision Transformers}: The use of AMP or 8-bit optimiser settings degrades the performance of both the transformer-based models to some extent in both datasets. This is in contrast to the CNN models, where there is no large performance degradation. One reason for the degradation in performance in transformers when using the 8-bit optimiser could be instability during training. The authors in~\cite{dettmers8} suggest the use of a {\tt stable embedding} layer when training transformers for NLP tasks. For a fairer comparison with the CNN models, we do not use the stable embedding layer for the two transformer models. Our experiments indicate that CNN models could be less sensitive when compared to transformer-based models in low/mixed precision or quantised settings.
\\
{\bf QALY-like metrics}: Quality Adjusted Life Year (QALY) is a utilitarian index used to estimate the cost-effectiveness of a health treatment; it attempts to arithmetically relate quality and quantity of life~\cite{prieto2003problems}.  In our experiments, we did not notice degradation in performance for some of the model classes. However, when there is a drop in performance, the tolerance levels in clinical settings would require formulating a Pareto optimisation of performance and resources~\cite{deb2011multi}. It is in this context that a metric similar to QALY could be useful, when balancing resource constraints with critical performance guarantees. 
\\
{\bf Limitations}: This work focused on optimising resource efficiency using only a few of the techniques (mixed precision training, half precision models, 8-bit optimisers). There are several other methods which could further optimise the resource consumption. For instance, formulating model selection that can yield inherently efficient models within neural architecture search (NAS) framework could be a possibility. However, performing NAS in itself is prohibitively expensive~\cite{elsken2019neural}. The other main limitation of the work is the gap in validating these methods in actual, resource constrained setting which the authors did not have access to. Testing these resource optimisation measures in diverse settings such as on different edge devices, for instance, could be an interesting direction for future work.

\vspace{-0.35cm}
\section{Conclusion}
\vspace{-0.35cm}
\label{sec:conc}
Large deep learning models are advancing medical image analysis and will undoubtedly accelerate clinical workflows. For these models to make the most impact, globally, resource efficiency is an important factor. In this work, we have evaluated a subset of resource optimisation methods that can reduce memory consumption, latency, and energy consumption, without any performance degradation (Densenet in RSNA, Swin Trans. in LIDC). In contrast to our initial hypothesis, surprisingly, there is no performance degradation across the board; only systematic ones (like with Transformers which can be handled). This prompts us to recommend these resource optimisation methods to be part of the standard procedures when developing and deploying deep learning methods.
\clearpage
\flushleft
{\bf Acknowledgments}
The authors would like to thank Christian Igel (UCPH) and Lee Lassen (Ambu A/S) for fruitful discussions, and Pedram Bakhtiarifard (UCPH) for their public repository for making radar plots. RS and JS are partly funded by European Union’s HorizonEurope research and innovation programme under grant agreements No. 101070284 and No. 101070408.

\bibliographystyle{splncs04}
\bibliography{references}

\begin{thebibliography}{10}
\providecommand{\url}[1]{\texttt{#1}}
\providecommand{\urlprefix}{URL }
\providecommand{\doi}[1]{https://doi.org/#1}

\bibitem{anthony2020carbontracker}
Anthony, L.F.W., Kanding, B., Selvan, R.: Carbontracker: Tracking and
  predicting the carbon footprint of training deep learning models. ICML
  Workshop on Challenges in Deploying and monitoring Machine Learning Systems
  (July 2020), arXiv:2007.03051

\bibitem{armato2004lung}
Armato~III, S.G., McLennan, G., McNitt-Gray, M.F., Meyer, C.R., Yankelevitz,
  D., Aberle, D.R., Henschke, C.I., Hoffman, E.A., Kazerooni, E.A., MacMahon,
  H., et~al.: Lung image database consortium: developing a resource for the
  medical imaging research community. Radiology  \textbf{232}(3),  739--748
  (2004)

\bibitem{bartoldson2022compute}
Bartoldson, B.R., Kailkhura, B., Blalock, D.: Compute-efficient deep learning:
  Algorithmic trends and opportunities. arXiv preprint arXiv:2210.06640  (2022)

\bibitem{baweja2018towards}
Baweja, C., Glocker, B., Kamnitsas, K.: Towards continual learning in medical
  imaging. Workshop on Medical Imaging meets NeurIPS (2018), arXiv:1811.02496

\bibitem{chakrabarti2019backprop}
Chakrabarti, A., Moseley, B.: Backprop with approximate activations for
  memory-efficient network training. Advances in Neural Information Processing
  Systems  \textbf{32} (2019)

\bibitem{cheng2020compression}
Cheng, Y., Wang, D., Zhou, P., Zhang, T.: Model compression and acceleration
  for deep neural networks: The principles, progress, and challenges. IEEE
  Signal Processing Magazine  \textbf{35}(1),  126--136 (2018)

\bibitem{rsna-breast-cancer-detection}
Chris~Carr, FelipeKitamura, H.k.i.J.K.C.J.M.K.A.M.V.M.R.R.B.S.D.: Rsna
  screening mammography breast cancer detection (2022),
  \url{https://kaggle.com/competitions/rsna-breast-cancer-detection}

\bibitem{daysal2022economic}
Daysal, N.M., Mullainathan, S., Obermeyer, Z., Sarkar, S.K., Trandafir, M.: An
  economic approach to machine learning in health policy. Univ. of Copenhagen
  Dept. of Economics Discussion, CEBI Working Paper (24) (2022)

\bibitem{deb2011multi}
Deb, K.: Multi-objective optimisation using evolutionary algorithms: an
  introduction. Springer (2011)

\bibitem{dettmers8}
Dettmers, T., Lewis, M., Shleifer, S., Zettlemoyer, L.: 8-bit optimizers via
  block-wise quantization. In: International Conference on Learning
  Representations (2022), \url{https://openreview.net/forum?id=shpkpVXzo3h}

\bibitem{dosovitskiyimage}
Dosovitskiy, A., Beyer, L., Kolesnikov, A., Weissenborn, D., Zhai, X.,
  Unterthiner, T., Dehghani, M., Minderer, M., Heigold, G., Gelly, S., et~al.:
  An image is worth 16x16 words: Transformers for image recognition at scale.
  In: International Conference on Learning Representations (2020)

\bibitem{elsken2019neural}
Elsken, T., Metzen, J.H., Hutter, F.: Neural architecture search: A survey. The
  Journal of Machine Learning Research  \textbf{20}(1),  1997--2017 (2019)

\bibitem{fiesler1990weight}
Fiesler, E., Choudry, A., Caulfield, H.J.: Weight discretization paradigm for
  optical neural networks. In: Optical interconnections and networks.
  vol.~1281, pp. 164--173. SPIE (1990)

\bibitem{huang2017densely}
Huang, G., Liu, Z., Van Der~Maaten, L., Weinberger, K.Q.: Densely connected
  convolutional networks. In: Proceedings of the IEEE conference on computer
  vision and pattern recognition. pp. 4700--4708 (2017)

\bibitem{hubara2017quantized}
Hubara, I., Courbariaux, M., Soudry, D., El-Yaniv, R., Bengio, Y.: Quantized
  neural networks: Training neural networks with low precision weights and
  activations. The Journal of Machine Learning Research  \textbf{18}(1),
  6869--6898 (2017)

\bibitem{kingma2014adam}
Kingma, D., Ba, J.: Adam optimizer. arXiv preprint arXiv:1412.6980 pp. 1--15
  (2014)

\bibitem{lidc}
Knegt, S.: A {P}robabilistic {U-N}et for segmentation of ambiguous images
  implemented in {P}y{T}orch.
  \url{https://github.com/stefanknegt/Probabilistic-Unet-Pytorch} (2018)

\bibitem{liu2021swin}
Liu, Z., Lin, Y., Cao, Y., Hu, H., Wei, Y., Zhang, Z., Lin, S., Guo, B.: Swin
  transformer: Hierarchical vision transformer using shifted windows. In:
  Proceedings of the IEEE/CVF international conference on computer vision. pp.
  10012--10022 (2021)

\bibitem{micikevicius2018mixed}
Micikevicius, P., Narang, S., Alben, J., Diamos, G., Elsen, E., Garcia, D.,
  Ginsburg, B., Houston, M., Kuchaiev, O., Venkatesh, G., Wu, H.: Mixed
  precision training. In: International Conference on Learning Representations
  (2018), \url{https://openreview.net/forum?id=r1gs9JgRZ}

\bibitem{nagel2021white}
Nagel, M., Fournarakis, M., Amjad, R.A., Bondarenko, Y., Van~Baalen, M.,
  Blankevoort, T.: A white paper on neural network quantization. arXiv preprint
  arXiv:2106.08295  (2021)

\bibitem{orel1999birads}
Orel, S.G., Kay, N., Reynolds, C., Sullivan, D.C.: Bi-rads categorization as a
  predictor of malignancy. Radiology  \textbf{211}(3),  845--850 (1999)

\bibitem{paszke2019pytorch}
Paszke, A., Gross, S., Massa, F., Lerer, A., Bradbury, J., Chanan, G., Killeen,
  T., Lin, Z., Gimelshein, N., Antiga, L., et~al.: Pytorch: An imperative
  style, high-performance deep learning library. In: Advances in Neural
  Information Processing Systems. pp. 8024--8035 (2019)

\bibitem{prieto2003problems}
Prieto, L., Sacrist{\'a}n, J.A.: Problems and solutions in calculating
  quality-adjusted life years (qalys). Health and quality of life outcomes
  \textbf{1}, ~1--8 (2003)

\bibitem{ricci2022addressing}
Ricci~Lara, M.A., Echeveste, R., Ferrante, E.: Addressing fairness in
  artificial intelligence for medical imaging. nature communications
  \textbf{13}(1), ~4581 (2022)

\bibitem{selvan2022carbon}
Selvan, R., Bhagwat, N., Wolff~Anthony, L.F., Kanding, B., Dam, E.B.: Carbon
  footprint of selecting and training deep learning models for medical image
  analysis. In: Medical Image Computing and Computer Assisted
  Intervention--MICCAI 2022: 25th International Conference, Singapore,
  September 18--22, 2022, Proceedings, Part V. pp. 506--516. Springer (2022)

\bibitem{sevilla2022compute}
Sevilla, J., Heim, L., Ho, A., Besiroglu, T., Hobbhahn, M., Villalobos, P.:
  Compute trends across three eras of machine learning. arXiv preprint
  arXiv:2202.05924  (2022)

\bibitem{sun2020ultra}
Sun, X., Wang, N., Chen, C.Y., Ni, J., Agrawal, A., Cui, X., Venkataramani, S.,
  El~Maghraoui, K., Srinivasan, V.V., Gopalakrishnan, K.: Ultra-low precision
  4-bit training of deep neural networks. Advances in Neural Information
  Processing Systems  \textbf{33},  1796--1807 (2020)

\bibitem{sze2017efficient}
Sze, V., Chen, Y.H., Yang, T.J., Emer, J.S.: Efficient processing of deep
  neural networks: A tutorial and survey. Proceedings of the IEEE
  \textbf{105}(12) (2017)

\bibitem{tan2019efficientnet}
Tan, M., Le, Q.: Efficientnet: Rethinking model scaling for convolutional
  neural networks. In: International conference on machine learning. pp.
  6105--6114 (2019)

\bibitem{sdg2022}
United-Nations: Sustainable development goals report (2022)

\bibitem{vokinger2021continual}
Vokinger, K.N., Feuerriegel, S., Kesselheim, A.S.: Continual learning in
  medical devices: Fda's action plan and beyond. The Lancet Digital Health
  \textbf{3}(6),  e337--e338 (2021)

\bibitem{yu2018artificial}
Yu, K.H., Beam, A.L., Kohane, I.S.: Artificial intelligence in healthcare.
  Nature biomedical engineering  \textbf{2}(10),  719--731 (2018)

\end{thebibliography}

\clearpage
\appendix
\section{Additional Results}

\begin{figure}[h]
\vspace{-1.0cm}
    \centering
    \includegraphics[width=0.69\textwidth]{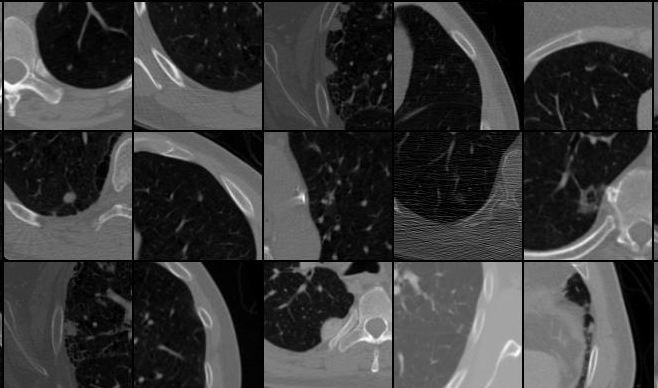}
    \caption{LIDC-IDRI dataset comprises 1018 thoracic CT images with lesions annotated by four radiologists~\cite{armato2004lung}. Patches of 128x128 px are extracted from the 2D slices, yielding a total of 15096 patches~\cite{lidc}. Each patch has annotations from four raters marking the tumour regions. These segmentation masks were converted into binary labels indicating the presence (if $\geq 2$ radiologists marked a tumour) or absence of tumours (if $<2$ raters marked tumours), resulting in a fairly balanced
dataset. All image intensities are normalised to be in [0, 1]. Training, validation and test splits are made following a [0.6,0.2,0.2] ratio.}
\vspace{-0.5cm}
\end{figure}
\vspace{-0.5cm}

\begin{table}[h]
\caption{Quantitative comparison of MLP and EfficientNet~\cite{sze2017efficient} reported over three random initialisations for the RSNA dataset. The use of 8-bit optimizer (8b\_Opt), automatic mixed precision (AMP) and half precision model (Half) are marked. Number of parameters: $|\theta|$, mean accuracy with standard deviation over three seeds, average GPU memory required (in GB), average inference time and average energy consumption of different settings reported when predicting on test set.}
\vspace{0.5cm}
\label{tab:app_results}
\centering
\scriptsize
\begin{tabular}{lrcccccrrrr}
\toprule
        {\bf Method} &   {\bf 
        $|\theta|$}(M) &  {\bf 8b\_Opt} &  {\bf AMP} &   {\bf Half} & {\bf Acc.}  &    {\bf GPU} &     {\bf T}$_{c}$(s) & {\bf T}(s) & {\bf E} (Wh) \\
\midrule

    \multirow{5}{*}{MLP} &   \multirow{5}{*}{8.4} &     \xmark &    \xmark & \xmark & 0.685$\pm$0.010 &   0.4 &   85.0 & 0.2 &  7.3 \\
            &&    \xmark &    \cmark & \xmark & 0.687$\pm$0.005  &     0.4 &   92.3 & 0.2&   8.1 \\
            &&    \cmark &    \xmark & \xmark & 0.693$\pm$0.007 &      0.2 &   41.3 & 0.2 & 3.4 \\
            &&    \cmark &    \cmark & \xmark & {\bf 0.694$\pm$0.005} &      0.2 &   55.3 & 0.2 & 4.7 \\
            &&    \cmark &    \xmark & \cmark & 0.680$\pm$0.005 &      0.1 &   27.3 & 0.2 & 2.4 \\
            \midrule
\multirow{5}{*}{Efficientnet\cite{tan2019efficientnet}} &   \multirow{5}{*}{51.1}&  \xmark &    \xmark & \xmark &  0.679$\pm$0.026 & 14.6 &  666.7 & 4.9 & 104.3 \\
 &&    \xmark &    \cmark & \xmark & 0.692$\pm$0.005  &  7.8 & 4027.3 & 4.9 & 618.4 \\
 &&    \cmark &    \xmark & \xmark & 0.715$\pm$0.022  & 14.2 &  466.0 & 4.9& 74.4 \\
 &&    \cmark &    \cmark & \xmark & {\bf 0.717$\pm$0.008}  &  7.4 & 2769.0 & 4.9& 427.2 \\ 
 &&    \cmark &    \xmark & \cmark & {\bf 0.717$\pm$0.008}  &  7.4 & 2769.0 & 3.8 & 427.2 \\ 
\bottomrule
\end{tabular}
\end{table}

\begin{figure}[h]
\centering
\begin{minipage}{0.3\textwidth}
    \centering
    \includegraphics[width=0.95\textwidth]{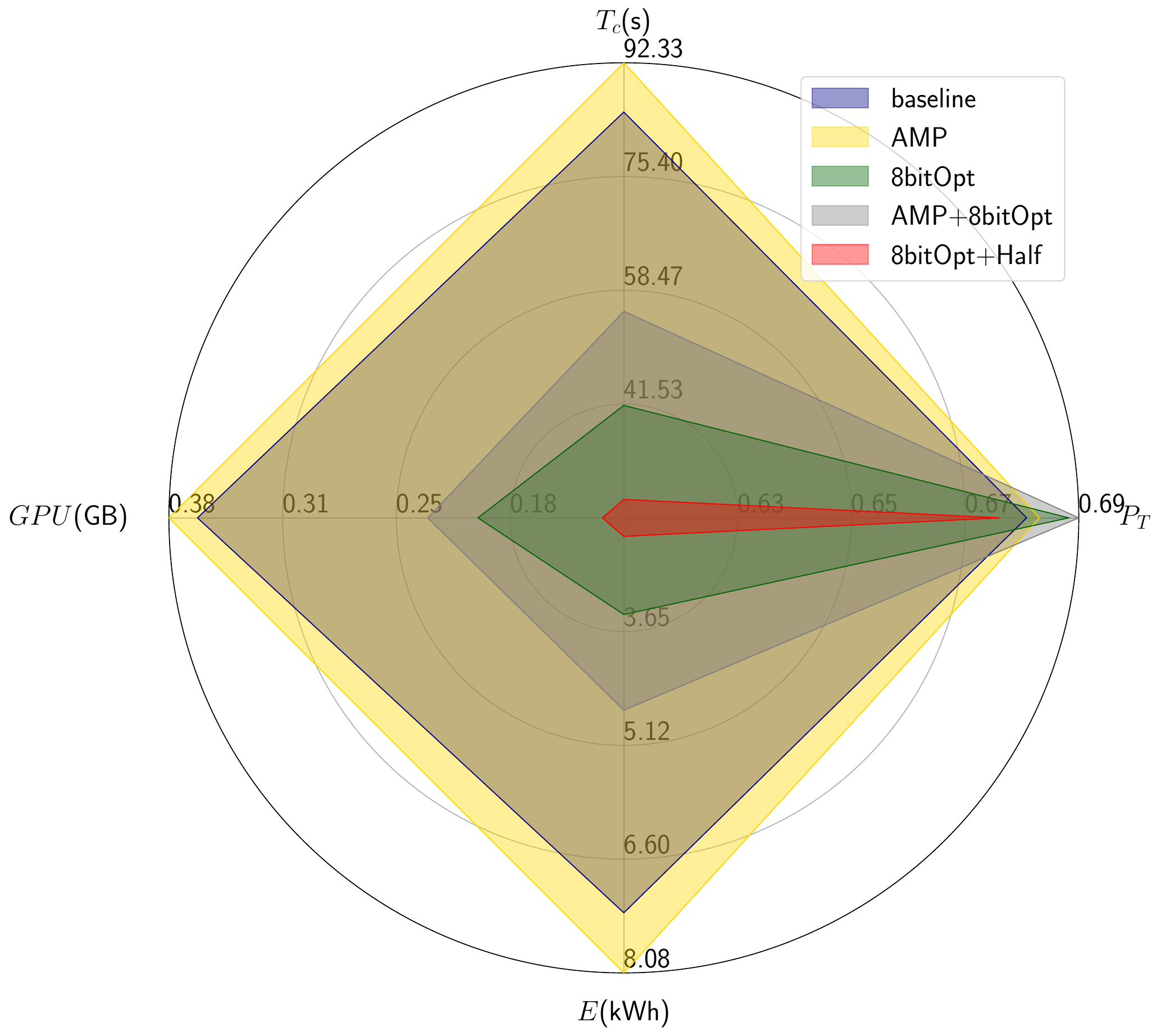}

(A): MLP
\end{minipage}
\begin{minipage}{0.3\textwidth}
    \centering
    \includegraphics[width=0.95\textwidth]{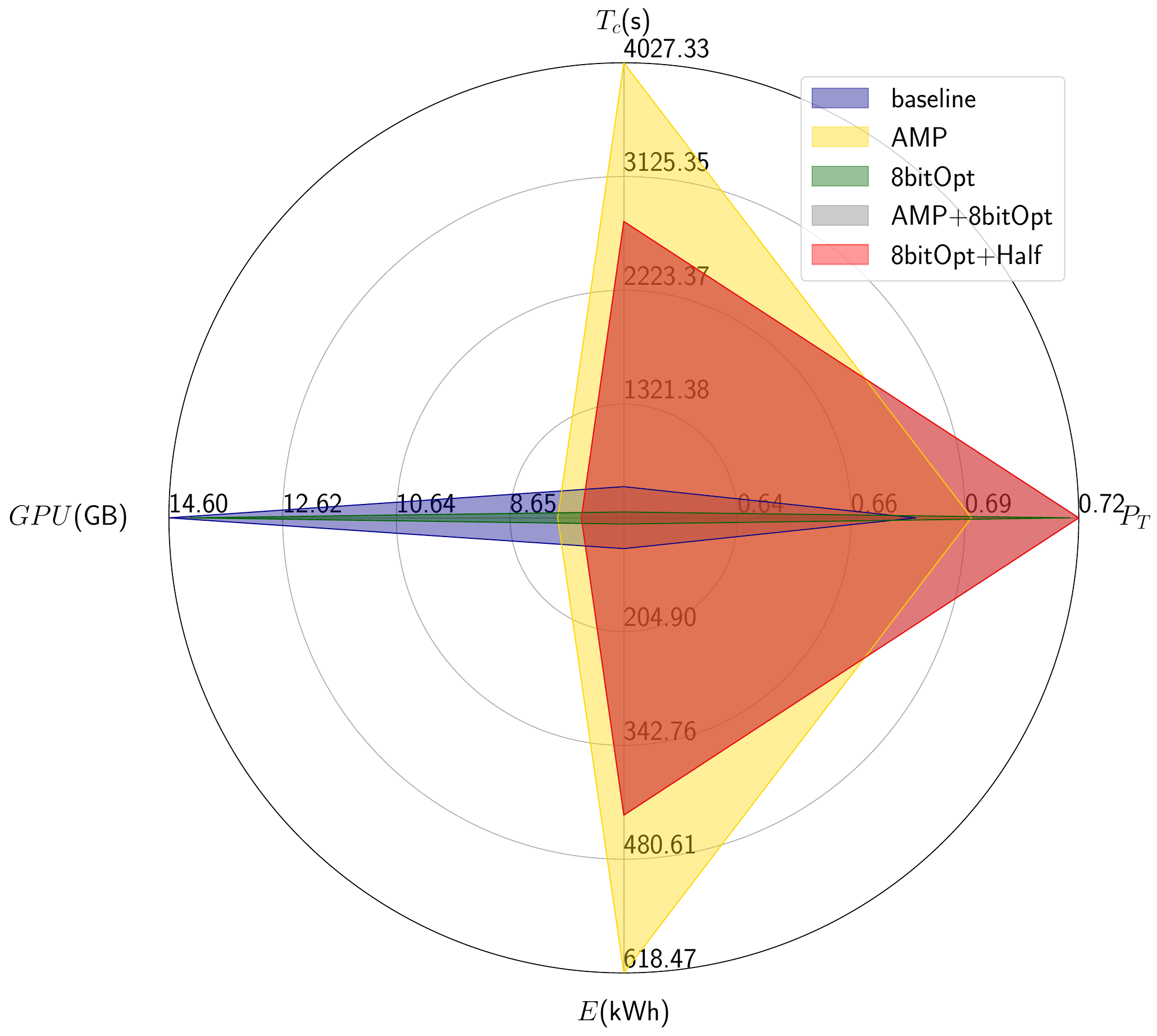}
    
    (B): EfficientNet~\cite{sze2017efficient}
\end{minipage}
    
\caption{(A \& B): Radar plots for multi-layered perceptron (MLP) and EfficientNet~\cite{sze2017efficient} showing the mean metrics for performance ($P_T, E, \text{GPU},T$) reported in Table~\ref{tab:app_results} for the five settings.}
    \label{fig:app_rsna}
\end{figure}


\begin{figure}
\begin{minipage}{0.3\textwidth}
    \centering
    \includegraphics[width=0.95\textwidth]{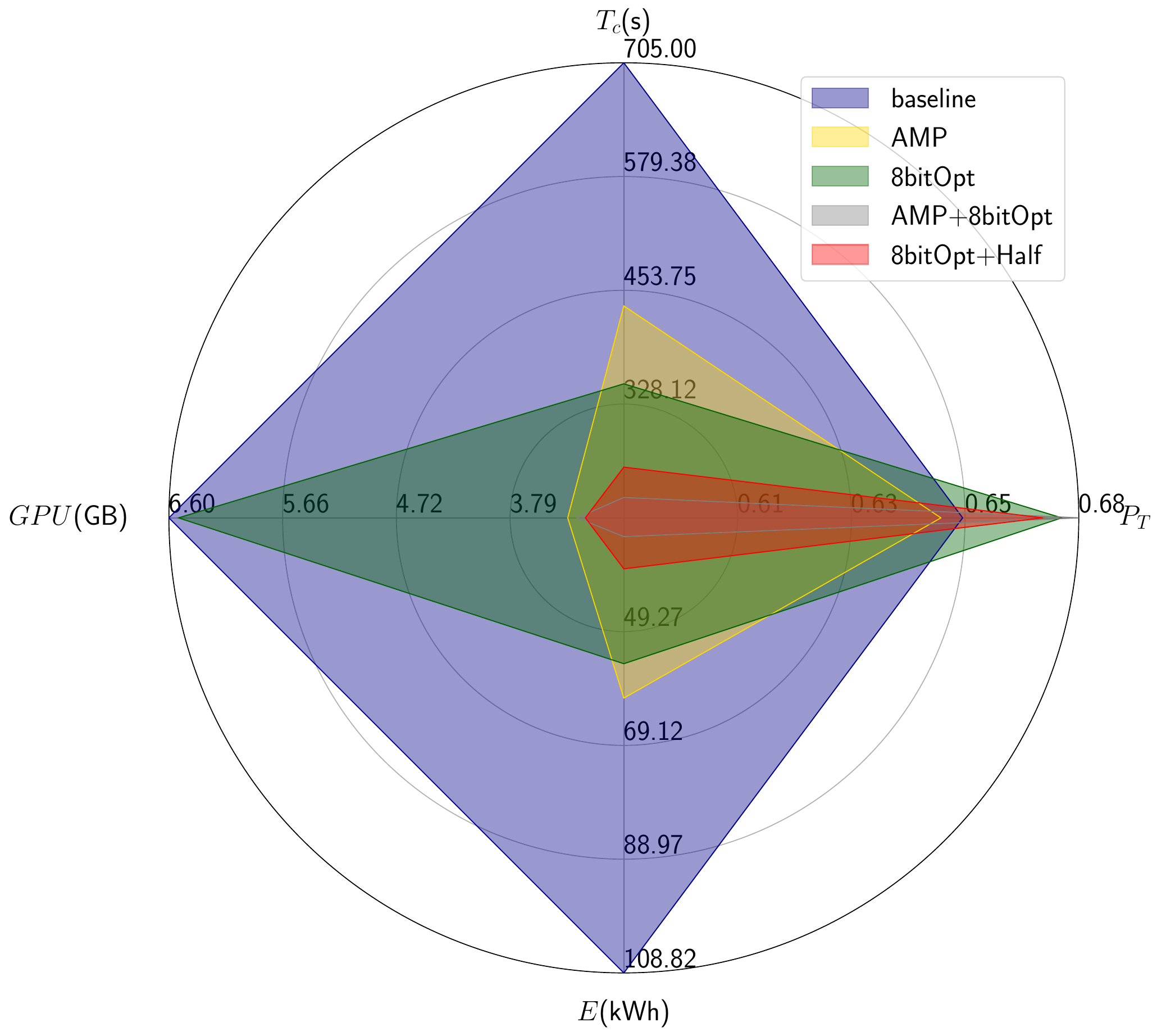}

(A)
\end{minipage}
\begin{minipage}{0.3\textwidth}
    \centering
    \includegraphics[width=0.95\textwidth]{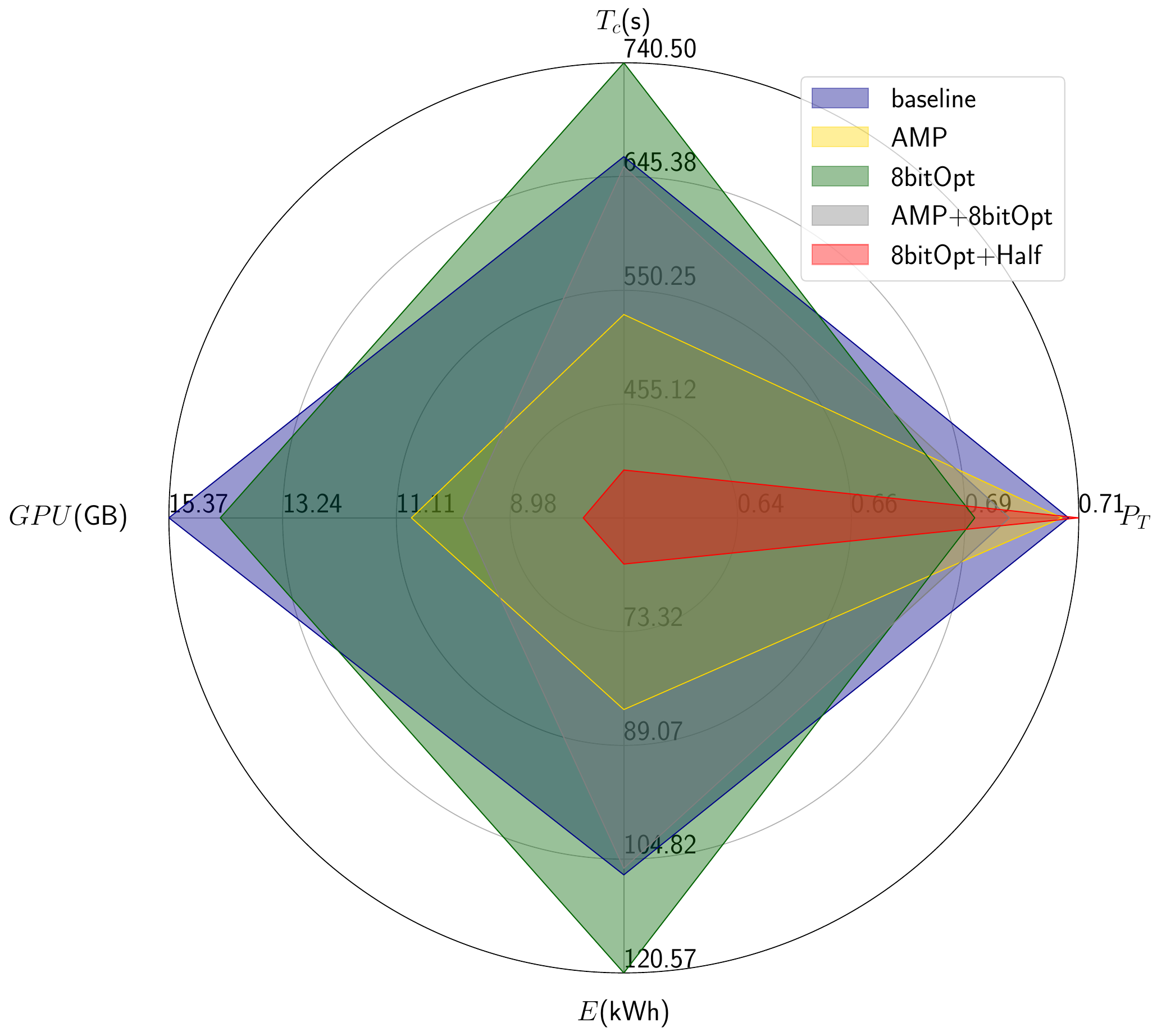}
    
    (B)
\end{minipage}
\begin{minipage}{0.3\textwidth}
    \centering
    \includegraphics[width=0.95\textwidth]{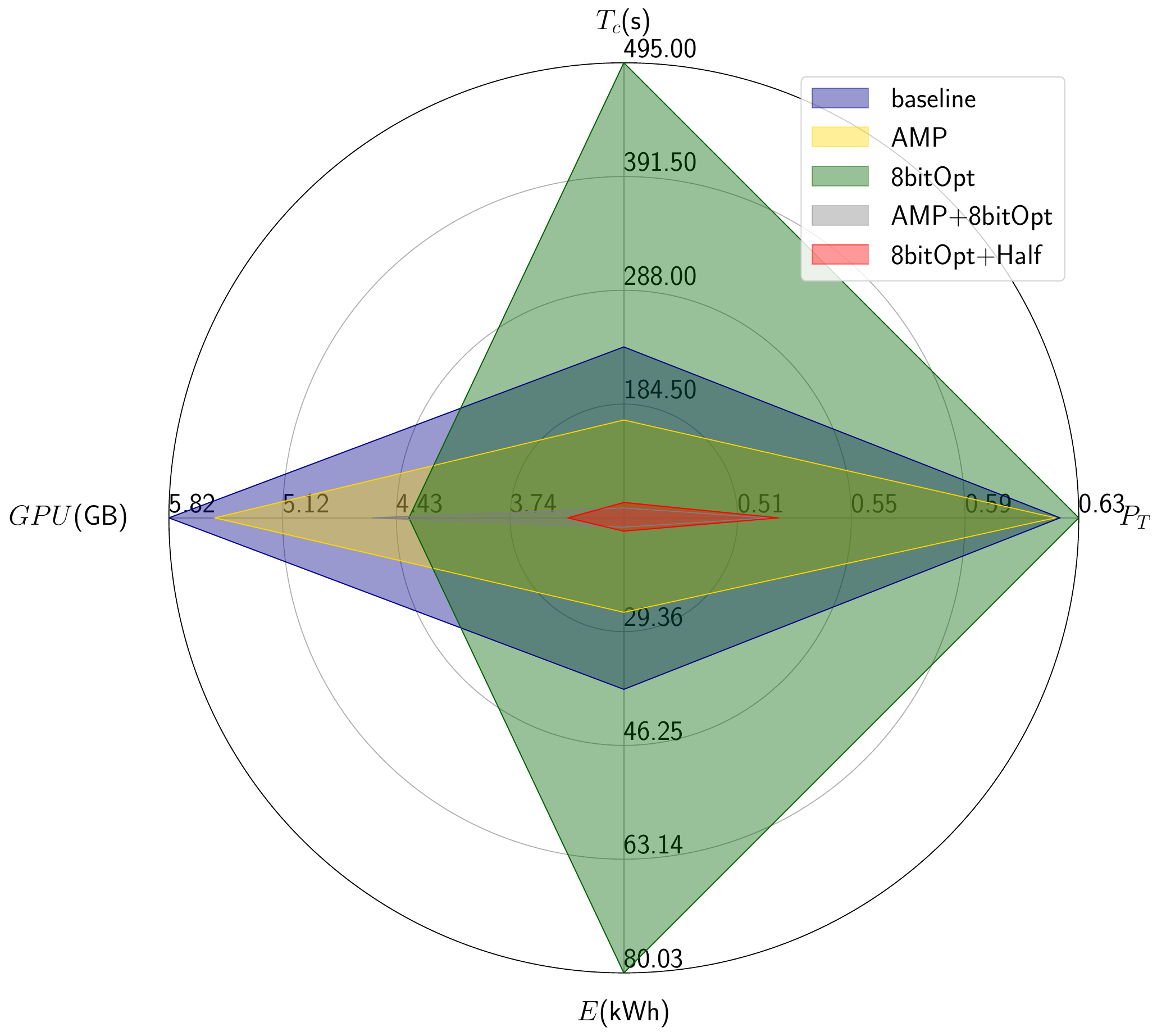}
    
    (C)
\end{minipage}

\caption{Radar plots for DenseNet~\cite{huang2017densely}, Swin Transformer~\cite{liu2021swin} and Vision Transformer~\cite{dosovitskiyimage} on the LIDC dataset, showing the mean metrics for performance ($P_T, E, \text{GPU},T$) reported in Table~\ref{tab:results} for the five settings.}

    \label{fig:app_lidc}
\end{figure}

\end{document}